\documentclass[journal]{IEEEtran}
\usepackage{latexsym,,amssymb,amsmath,graphicx,epsf,cite,bbm,float}
\usepackage{ifpdf}
\usepackage{epstopdf}

\usepackage{algorithm,algorithmic}
\usepackage{amsmath,amssymb,bm}
\usepackage{amsfonts,dsfont,color,bbm,subcaption}

\newcommand{\abs}[1]{|#1|}

\DeclareMathOperator*{\argmin}{arg\,min}

\usepackage{hyperref,amsthm}

\newtheorem{theorem}{Theorem}

\newtheorem{lemma}[]{Lemma}

\newtheorem{proposition}[]{Proposition}

\newtheorem{corollary}[]{Corollary}

\newtheorem{remark}[]{Remark}

\newtheorem{definition}[]{Definition}

\newtheorem{example}[]{Example}

\begin{document}

\title{Nonconvex Extension of Generalized Huber Loss for Robust Learning and Pseudo-Mode Statistics} 
\author{\IEEEauthorblockN{Kaan Gokcesu}, \IEEEauthorblockN{Hakan Gokcesu} }
\maketitle

\begin{abstract}
	We propose an extended generalization of the pseudo Huber loss formulation. We show that using the log-exp transform together with the logistic function, we can create a loss which combines the desirable properties of the strictly convex losses with robust loss functions. With this formulation, we show that a linear convergence algorithm can be utilized to find a minimizer.
	We further discuss the creation of a quasi-convex composite loss and provide a derivative-free exponential convergence rate algorithm.
\end{abstract}

\section{Introduction}

In the fields of statistics, decision theory, learning and optimization \cite{poor_book,cesa_book,huberbook,portnoy2000robust}, it has become paramount to design robust decision makers, where a trained or learned model is minimally influenced by some outlying anomalies in comparison with the inlying nominal data  \cite{hastie2019statistical,huber2004robust}. 
Especially in the tasks of parameter estimation (or learning in general), it has become popular to use robust loss functions because of their desirability in comparison with the more traditional loss functions that are very sensitive to large errors. 

To achieve robustness, certain parametric regularity formulations are studied \cite{barron2019general}; which has been applied in early vision \cite{black1996unification} and conic fitting \cite{zhang1997parameter}. However, the popular approach is to achieve robustness by using suitable loss functions \cite{rosasco2004loss}.
When using gradient descent or M-estimation\cite{hampel2011robust}, a variety of losses are tried empirically to design a well performing learning system.  Although, in some applications, the loss function is inherent to the problem itself, a carefully designed loss metric can be substantially helpful in the performance evaluation of the learning algorithms; specifically, in the problems of parameter estimation \cite{gokcesu2018density,beck1977parameter} and sequential prediction \cite{gokcesu2016prediction,singer,neyshabouri2018asymptotically}. 
It has become crucial to achieve robustness intrinsically with the help of well designed loss functions instead of external approaches like anomaly detection methods \cite{rousseeuw2005robust,gokcesu2018anomaly,rousseeuw2011robust,gokcesu2019outlier}.

In optimization problems, there are three traditional loss functions and their consequent centralizing statistics:
\begin{enumerate}
	\item L2-loss: $L(x)=x^{2}$ (square loss) together with its minimizer, the mean.
	\item L1-loss: $L(x)=|x|$ (absolute loss) together with its minimizer, the median.
	\item L0-loss: $L(x)=1-\delta(x)$ (Hamming loss, where $\delta(0)=1$ and $0$ elsewhere) and its minimizer, the mode.
\end{enumerate}
Since the square loss is strongly convex, it has fast learning performance. However, because of its sensitivity, it is prone to be erroneous in the presence of large outliers \cite{huber2004robust}. In comparison, the absolute loss is robust against such outliers but has slower learning capabilities. Even though Hamming loss has minimal sensitivity to outliers, its use case is limited in high precision datasets.

To this end, it is paramount to combine the desirable qualities of different loss functions, for the sake of efficient and robust learning.
The trivial approach is to use a piecewise combination, e.g.,
\begin{align}
	L_P(x)=\begin{cases} 
		x^2 &, \abs{x}\leq \sigma_1\\
		\frac{1-\sigma_1^2}{\sigma_2-\sigma_1}\abs{x}+\frac{\sigma_1^2\sigma_2-\sigma_1}{\sigma_2-\sigma_1} &, \sigma_2\geq\abs{x}>\sigma_1\\
		1&, \abs{x}>\sigma_2.
	\end{cases}.
\end{align}
In this formulation, we have the requirement that $\sigma_1^2\leq 1$. To relax this, we can parametrize the formulation as follows:
\begin{align}
	L_P(x)=\begin{cases} 
		\alpha x^2 &, \abs{x}\leq \sigma_1\\
		\frac{1-\alpha\sigma_1^2}{\sigma_2-\sigma_1}\abs{x}+\frac{\alpha\sigma_1^2\sigma_2-\sigma_1}{\sigma_2-\sigma_1} &, \sigma_2\geq\abs{x}>\sigma_1\\
		1&, \abs{x}>\sigma_2
	\end{cases},
\end{align}
where the requirement is relaxed to $\alpha\sigma_1^2\leq 1$ and $\alpha$ is a free parameter. Note that this loss is not differentiable and formulations like Huber loss \cite{huberbook} are needed for differentiability. However, not even those formulations are smooth.

To achieve smoothness, alternative formulations have been proposed for the combination of square and absolute losses \cite{lange1990convergence}. The most popular one is Pseudo-Huber loss \cite{charbonnier1997deterministic}, i.e.,
\begin{align}
L_{PH}(x)=\delta\sqrt{1+\frac{x^2}{\delta^2}}.
\end{align}
The work in \cite{gokcesu2021generalized}, provides a Generalized Huber Loss smoothing, where the most prominent convex example is
\begin{align}
	L_{GH}(x)=\frac{1}{\alpha}\log(e^{\alpha x}+e^{-\alpha x}+\beta),
\end{align}
which is the log-cosh loss when $\beta=0$ \cite{logcosh}.

Hence, to create smooth approximations for the combination of strongly convex and robust loss functions, the popular approach is to utilize the Huber loss or its variants. Even though the square and absolute losses are smoothly combined in various forms in literature; the incorporation of Hamming loss, and bounded losses in general, is lacking. To this end, we propose an extended formulation for the Generalized Huber loss (where the asymptotes can be more freely designed), which encompasses many of its predecessor formulations. We also propose derivative-free approaches to obtain minimizers in an efficient way. 

The organization of our paper is as follows. In \autoref{sec:huber}, we provide the extended generalization of the smooth Huber loss. In \autoref{sec:nonconvex}, we design a smooth and robust loss function together with its non-convex solver. In \autoref{sec:quasi}, we discuss how to produce a quasi-convex composite loss with our design and provide a derivative-free solver with exponential convergence rate. In \autoref{sec:conc}, we finish with further discussions and concluding remarks. 

\section{Extended Generalized-Huber Loss}\label{sec:huber}
In this section, we extend the Generalized-Huber loss definition in \cite{gokcesu2021generalized}.
A loss function is expected to have some desirable properties.

\begin{definition}\label{def:L}
	For a loss function $L(\cdot)$ such that
	\begin{align*}
		L:\Re\rightarrow\Re,
	\end{align*}
	we desire the following:
	\begin{enumerate}
		\item $L(\cdot)$ has a minimum at $x=0$, i.e.,
		\begin{align*}
			\min_{x\in\Re}L(x)=L(0).
		\end{align*}
		
		\item At $x=0$, $L(\cdot)$ has a positive second derivative, i.e.,
		\begin{align*}
			L''(0)>0,
		\end{align*}
		and finite higher derivatives for convergence to quadratic function near $0$ from Taylor's expansion, i.e.,
		\begin{align*}
			L(x)\approxeq L(0)+\frac{1}{2}L''(0)x^2.
		\end{align*}
		
		\item The loss function $L(\cdot)$ is quasi-convex, i.e.,
		\begin{align*}
			L(x_1)\leq L(x_2)&&\forall x_1,x_2: 0\leq x_1\leq x_2,\\
			L(x_1)\leq L(x_2)&&\forall x_1,x_2: 0\geq x_1\geq x_2.
		\end{align*} 
		Hence, $L(\cdot)$ is nonincreasing for $x\leq 0$ and nondecreasing for $x\geq0$.
	\end{enumerate}
\end{definition}

\autoref{def:L} covers intuitive and nice properties for general loss functions. To design such a loss, we utilize a monotone nondecreasing auxiliary function $f(\cdot)$ similar  to \cite{gokcesu2021generalized}.
\begin{definition}\label{def:f}
	The auxiliary function $f(x)$ needs to satisfy the following properties:
	\begin{enumerate}
		\item 	$\lim_{x\rightarrow\infty}f(x)=\infty,$
		\item $		f(x)<\infty, \forall x<\infty.$
		\item $f(x)$ is convex
	\end{enumerate}
\end{definition}
Hence, $f(\cdot)$ is a convex function that is divergent towards infinity and convergent otherwise. Using this auxiliary function, the extended generalized Huber loss is defined as follows. 
\begin{definition}\label{def:g}
	The loss function is given by
	\begin{align*}
		L_E(x)=g(f(x)+f(-x)),
	\end{align*}
	where $g(\cdot)$ is a suitable monotone increasing transform.
\end{definition}
Unlike \cite{gokcesu2021generalized}, the function $g(\cdot)$ is a suitable transform not limited to the inverse of $f(\cdot)$. Given that the functions $f(\cdot)$ and $g(\cdot)$ are smooth and differentiable, the loss function $L_E(\cdot)$ is also smooth and differentiable.

\begin{lemma}
	$L_E(x)$ has a global minimizer at $x=0$.
	\begin{proof}
		From \autoref{def:g}, we have
		\begin{align*}
			\argmin_x L_E(x)&=\argmin_x g(f(x)+f(-x)),\\
			&=\argmin_x [f(x)+f(-x)],
		\end{align*}
		which is $0$ since $g(\cdot)$ is a monotone increasing function and $f(\cdot)$ is convex.
	\end{proof}
\end{lemma}

\begin{lemma}\label{thm:asymptotes}
	The loss function $L_E(\cdot)$ has the following asymptotic behavior:
	\begin{align*}
		\lim_{\abs{x}\rightarrow\infty}L_E(x)=g\circ f(\abs{x})
	\end{align*}
	\begin{proof}
		From \autoref{def:f}; when $x$ goes to $\infty$, only $f(x)$ is divergent and when $x$ goes to $-\infty$, only $f(-x)$ is divergent. Hence,
		\begin{align}
			&L_E(x)\rightarrow g\circ f(x) &&\text{ as } {x\rightarrow\infty},\\
			&L_E(x)\rightarrow g\circ f(-x) &&\text{ as }{x\rightarrow-\infty},
		\end{align}
		which concludes the proof.
	\end{proof}
\end{lemma}

\begin{lemma}\label{thm:nearzero}
	$L_E(x)$ converges to the following quadratic loss near $0$:
	\begin{align*}
		L_E(x)\rightarrow g'(2f(0))f''(0)x^2+g(2f(0)) &&\text{ as }\abs{x}\rightarrow 0.
	\end{align*}
	\begin{proof}
		The proof comes from \cite{gokcesu2021generalized}.
	\end{proof}
\end{lemma}
\begin{remark}
	Since $g'(\cdot)>0$ from \autoref{def:g}, we have
	\begin{align*}
		L_E''(0)>0 \iff f''(0)>0.
	\end{align*}
\end{remark}

\begin{example}
	When the transform function $g(\cdot)$ is selected as the inverse (or pseudo-inverse) of $f(\cdot)$; we have the design in \cite{gokcesu2021generalized}, where
	\begin{align*}
		L_E(x)\rightarrow ax^2+b && \text{as }\abs{x}\rightarrow0,
	\end{align*}
	for some $a>0, b\in\Re$;	and
	\begin{align*}
		L_E(x)\rightarrow f^{-1}\circ f (\abs{x})=\abs{x} && \text{as }\abs{x}\rightarrow\infty.
	\end{align*}
\end{example}

\begin{example}
	When the transform function $g(\cdot)$ is selected as the composite of some monotone increasing function $h(\cdot)$ with $f^{-1}(\cdot)$ (inverse or pseudo-inverse of $f(\cdot)$), i.e.,
	\begin{align*}
		g(\cdot)=h\circ f^{-1}(\cdot),
	\end{align*}
	we have again a quadratic behavior near $0$ and the following asymptotic behavior:
	\begin{align*}
		L_E(x)\rightarrow h\circ f^{-1}\circ f (\abs{x})=h(\abs{x}) && \text{as }\abs{x}\rightarrow\infty.
	\end{align*}
	When $h(\cdot)$ is selected as $h(x)=x$, we have the result of \cite{gokcesu2021generalized}. If we select $h(x)=\sqrt{x}$, we get
	\begin{align*}
		L_E(x)\rightarrow h\circ f^{-1}\circ f (\abs{x})=\sqrt{\abs{x}} && \text{as }\abs{x}\rightarrow\infty.
	\end{align*}
	In general, we can select $h(\cdot)$ as any concave asymptote tailored in accordance with the problem specifics.
\end{example}

\begin{remark}
	We can also dissect the loss function definition into its two asymptotes such that the function is given by
	\begin{align*}
		L_E(x)=g\circ f(x)+g\circ f(-x).
	\end{align*}
From \autoref{def:f} and \autoref{def:g}, we observe that the asymptotic behaviors remain the same, i.e.,
\begin{align*}
	L_E(x)\rightarrow g\circ f(\abs{x})&&\text{as }\abs{x}\rightarrow\infty.
\end{align*}
For a global minimum at $x=0$ (quasi-convexity), we need 
\begin{align*}
	L_E'(x)=&g'(f(x))f'(x)-g'(f(-x))f'(-x)\geq 0 \text{ for }x\geq 0. 
\end{align*}
For quadratic behavior near $0$, we need positive second derivative at $x=0$ from \autoref{def:L}, i.e.,
\begin{align*}
	L_E''(0)=&2g''(f(0))f'^2(0)+2g'(f(0))f''(0)>0.
\end{align*}
\end{remark}

\section{Derivative-Free Optimization of Non-convex Smoothed Hamming Loss}\label{sec:nonconvex}
Huber loss is proposed as a way to combine the fast optimization of square loss with the robust absolute loss. Hence, these types of loss functions provide some kind of statistic that is between the mean and the median, or in a sense a pseudo-median. However, if we design the asymptotes differently, we can extend this application for different statistics such as the mode. Normally, the mode of a sample set can be found by optimizing the cumulative Hamming distance (L0-loss), where the loss is an inverted delta function, which is $0$ at zero and 1 at everywhere else. However, we can design a smoothed Hamming loss using the framework in \autoref{sec:huber}.
\subsection{Non-convex Smoothed Hamming Loss}
To this end, we can utilize the class of sigmoid functions as the monotone transform function $h(\cdot)$. As in \cite{gokcesu2021generalized}, we utilize the exponential transform for the loss design, where
\begin{align}
	f(x)=&e^{ax}+b,\\	f^{-1}(x)=&\frac{1}{a}\log(x-b), &&x>b,\\
	g(\cdot)=&h\circ f^{-1}(\cdot),
\end{align}
for $b>-2$. We choose $h(\cdot)$ as the most commonly used sigmoid, a logistic function, i.e.,
\begin{align}
	h(x)=\frac{d}{c+e^{-kx}},
\end{align}
for $c,d,k>0$. Hence, the loss is given by
\begin{align}
	L_H(x)=&h\circ f^{-1}(f(x)+f(-x)),\\
	=&h\left(\frac{1}{a}\log(e^{ax}+e^{-ax}+b)\right).
\end{align}
Setting $k=a$, we get
\begin{align}
	L_H(x)=&\frac{d}{c+\frac{1}{e^{ax}+e^{-ax}+b}}.
\end{align}

\begin{lemma}
	The equivalent loss has the following form:
	\begin{align*}
		L_H(x)=-\frac{1}{e^{kx}+e^{-kx}+m},
	\end{align*}
	for some scaling parameter $k>0$ and smoothing parameter $m>-2$. 
	\begin{proof}
		The optimization of any affine transform of the loss function is equivalent because of linearity. Hence, $d$ is redundant and can be set to $c$. Thus,
		\begin{align}
			L_H(x)
			=&1-\frac{1}{c(e^{ax}+e^{-ax}+b)+1}
		\end{align}
		Again using affine transform, the loss function takes the final form after some substitutions.
	\end{proof}
\end{lemma}

\begin{remark}\label{thm:region}
	This loss function has three operating regions for $x\geq0$ (and their origin symmetries for $x\leq 0$). They are:
	\begin{enumerate}
		\item Convex region: $L_H''(x)\geq 0$
		\item Concave region: $L_H''(x)\leq 0$ and $xL_H'''(x)\leq 0$
		\item Tail region: $L_H''(x)\leq 0$ and $xL_H'''(x)\geq 0$
	\end{enumerate}
\end{remark}

\begin{theorem}
	When $m=2$, the loss is divided uniformly (in equal parts) between the three distinct regions in \autoref{thm:region}.
	\begin{proof}
		Let $\alpha(x)=e^{kx}+e^{-kx}$.
		We have the following critical points for the derivatives:
		\begin{itemize}
			\item $			L_H'(x)=0\implies \alpha(x)=2,$
			\item $L_H''(x)=0\implies 8+m\alpha(x)-\alpha^2(x)=0,$
			\item$
			L_H'''(x)=0\implies (\alpha^2(x)-4m\alpha(x)+m^2-24)\alpha'(x)=0.$
		\end{itemize}
		Since $m=2$, we have: 
		\begin{itemize}
			\item $L_H'(x)=0,\enspace \alpha(x)\in\{2\},$
			\item $L_H''(x)=0,\enspace\alpha(x)\in\{-2,4\},$
			\item
			$L_H'''(x)=0,\enspace\alpha(x)\in\{-2,2,10\}.$
		\end{itemize}
		Consequently, the corresponding three operating regions are as follows (since $\alpha(x)\geq 2$):
		\begin{enumerate}
			\item Convex region: $\alpha(x)\in[2,4)$, $L_H(x)\in[-\frac{1}{4},-\frac{1}{6})$,
			\item Concave region: $\alpha(x)\in[4,10)$, $L_H(x)\in[-\frac{1}{6},-\frac{1}{12})$,
			\item Tail region: $\alpha(x)\in[10,\infty)$, $L_H(x)\in[-\frac{1}{12},0)$,
		\end{enumerate} 
		which partitions the loss uniformly.
	\end{proof}
\end{theorem}

\subsection{Derivative-Free Non-convex Optimization}
For equivariance under translation and scaling, we normalize the dataset to the convex set $[0,1]$.
Given a set of samples ${x}_1,\ldots, {x}_N \in\{0,1\}$; we have the following objective function, i.e., average loss
\begin{align}
	\min_{{x}\in\Re}\frac{1}{N}\sum_{n=1}^{N}L_H({x}-{x}_n).\label{eq:obj}
\end{align}
The loss function $L_H(x)$ has maximum first derivative at the boundary of the convex region $\alpha(x)=4$. When $\alpha(x)=4$, $L'_H(x)<\frac{k}{9}$ . Thus, the average loss is Lipschitz continuous with $\frac{k}{9}$; and we can straightforwardly utilize the univariate global optimization algorithms in \cite{gokcesu2022low}.


Using $x_n$, the algorithm works as the following.
\begin{enumerate}
	\item At the start, we sample the boundaries $x=0$, $x=1$; and receive their evaluations $C(0)$, $C(1)$; where $C(\cdot)$ is the cumulative objective function in \eqref{eq:obj}. 
	\item Inputting $x_0=0, x_1=1, C_0=C(0), C_1=C(1)$; we determine the query $x'=\frac{x_0+x_1}{2}$ with its score $s'=\min(C(x_0),C(x_1))-\frac{k}{9}\left|\frac{x_0-x_1}{2}\right|$; and add to the list. \label{item:candidate}
	\item We sample the query with the lowest score from the query list and remove it. Let the sampled query be $x_m$ and its evaluation $C_m=C(x_m)$. Let $x_m$ be between the previous queries $x_l$ and $x_r$ with the corresponding evaluations $C(x_l)=C_l$ and $C(x_r)=C_r$ respectively. \label{item:sample}
	\item We repeat Step \ref{item:candidate} with the inputs: $x_0=x_l$, $x_1=x_m$, $C_0=C(x_l)$, $C_1=C(x_m)$.
	\item We repeat Step \ref{item:candidate} with the inputs: $x_0=x_m$, $x_1=x_r$, $C_0=C(x_m)$, $C_1=C(x_r)$.
	\item We return to Step \ref{item:sample}.
\end{enumerate}

\begin{remark}
	From \cite{gokcesu2022low}, we can achieve $\epsilon$-closeness to the optimal loss in $O(k\epsilon^{-1})$ evaluations. Since each evaluation takes $O(N)$ time, our computational complexity is $O(Nk\epsilon^{-1})$.
\end{remark}

\begin{remark}
	Although choosing $k$ large will approximate the Hamming loss better, it will also increase the Lipschitz continuity parameter, which decreases the convergence performance of the algorithm.
\end{remark}

\section{Derivative-Free Optimization of Quasi-Convex Smoothed Hamming Loss}\label{sec:quasi}
In the previous section, an approximate $\epsilon$-close solution, i.e.,
\begin{align}
	\left|\frac{1}{N}\sum_{n=1}^{N}L_H(\hat{x}-x_n)-\min_{x\in\Re}\frac{1}{N}\sum_{n=1}^{N}L_H({x}-x_n)\right|\leq \epsilon
\end{align} is found in a number of evaluations that is reciprocally dependent on $\epsilon$, i.e., closeness to the optimal loss, since the loss is non-convex. To this end, for increased efficiency, we can utilize the free scaling parameter $k$ to design a quasi-convex cumulative objective function.

\subsection{Preliminaries}
\begin{remark}
	For any loss function $L(\cdot)$, we achieve convexity when its second derivative is always nonnegative, i.e.,
	\begin{align*}
		L''(x)\geq 0, &&x\in\Re.
	\end{align*}
	Quasi-convexity has a weaker regularity condition, where given the optimal point $x_*$, we need
	\begin{align*}
		L(x_1)\leq& L_(x_2), &&x_*\leq x_1\leq x_2,\\
		L(x_1)\leq& L_(x_2), &&x_*\geq x_1\geq x_2.
	\end{align*}
\end{remark}

Note that even though quasi-convex loss functions are not simple to analyze, they have the following nice property.  
\begin{lemma}
	$K\circ L(\cdot)$ is quasi-convex when $L(\cdot)$ is quasi-convex and $K(\cdot)$ is monotone nondecreasing.
	\begin{proof}
		From quasi-convexity of $L(\cdot)$, we have
		\begin{align}
			L(\lambda x+(1-\lambda)y)\leq \max(L(x),L(y)).
		\end{align}
		Since $K(\cdot)$ is nondecreasing, we have
		\begin{align}
			K\circ L(\lambda x+(1-\lambda)y)\leq& K(\max(L(x),L(y))),\\
			\leq&\max (K\circ L(x), K\circ L(y)), 
		\end{align}
		which concludes the proof.
	\end{proof}
\end{lemma} 

\begin{lemma}\label{thm:quasi}
	$L(\cdot)$ is quasi-convex if $L(\cdot)$ has a lower bounded second derivative, which is also nonnegative wherever the absolute of the first derivative is small, i.e.,
	\begin{align*}
		L''(x)\geq& H, &&x\in\Re\\
		L''(x)\geq& 0, &&x\in\mathcal{X}=\{x:|L'(x)|\leq\delta\},
	\end{align*}
	for some $H<0$ and $\delta>0$.
	\begin{proof}
		Let us define the following function
		\begin{align}
			Q(x)=e^{\lambda L(x)},
		\end{align}
		where $\lambda>0$. $Q(\cdot)$ has the following derivatives:
		\begin{align}
			Q'(x)=&\lambda L'(x) e^{\lambda L(x)},\\
			Q''(x)=&\lambda L''(x) e^{\lambda L(x)}+\lambda^2 L'^2(x) e^{\lambda L(x)},\\
			Q''(x)=&\left( L''(x)+\lambda L'^2(x)\right)\lambda Q(x).
		\end{align}
		$Q(x)$ and $\lambda$ are positive and $L''(x)$ is nonnegative wherever $|L'(x)|\leq\delta$. Thus, for sufficiently large $\lambda$, we have $Q''(x)\geq 0$, i.e., $Q(\cdot)$ is convex. Since every convex function is also quasi-convex and $\log(\cdot)$ is monotone nondecreasing; $L(\cdot)$ is also quasi-convex, which concludes the proof.
	\end{proof} 
\end{lemma}

\subsection{Non-convex Smoothed Hamming Loss}
\begin{definition}\label{def:alpha}
	Let the cumulative objective function be
	\begin{align*}
		C(x)=\sum_{n=1}^{N}-\frac{1}{\alpha_n(x)+2}.,
	\end{align*}
	where $\alpha_n(x)=e^{k(x-x_n)}+e^{-k(x-x_n)}$. 
\end{definition}
\begin{lemma}\label{thm:alpha'}
	We have
	\begin{align*}
		\alpha'^2_n(x)=&k^2(\alpha_n^2(x)-4),\\
		\alpha''_n(x)=&k^2\alpha_n(x),
	\end{align*}
	when $\alpha_n(x)=e^{k(x-x_n)}+e^{-k(x-x_n)}$ is as in \autoref{def:alpha}.
	\begin{proof}
		The proof is straightforward from the first and second derivatives of $\alpha_n(x)$.
	\end{proof}
\end{lemma}

\begin{lemma}
	For the objective function $C(x)$ in \autoref{def:alpha}, we have
	\begin{align*}
		C'(x)=&\sum_{n=1}^{N}\frac{\alpha_n'(x)}{(\alpha_n+2)^2},\\
		C''(x)=&k^2\sum_{n=1}^{N}\frac{4-\alpha_n(x)}{(\alpha_n(x)+2)^2}.
	\end{align*}
	\begin{proof}
		The proof comes from utilizing \autoref{thm:alpha'} in the derivatives of $C(x)$.
	\end{proof}
\end{lemma}

\begin{definition}\label{def:p}
	Let us define the following probabilities
	\begin{align*}
		p_x(n)=\frac{(\alpha_n(x)+2)^{-2}}{Z_x},
	\end{align*}
	for $n\in\{1,\ldots,N\}$, where
	\begin{align*}
		Z_x=\sum_{n=1}^{N}(\alpha_n(x)+2)^{-2}.
	\end{align*}
\end{definition}

\begin{corollary}
	Using \autoref{def:p}, we have the following alternative expressions:
	\begin{align*}
		C'(x)=&Z_x\mathbb{E}_{p_x}[\alpha_n'(x)],\\
		C''(x)=&k^2Z_x\mathbb{E}_{p_x}[4-\alpha_n(x)],
	\end{align*}	
	where $\mathbb{E}_{p_x}$ is the expectation over $p_x(n)$ probabilities.
\end{corollary}

\begin{proposition}
	If $k$ is nonzero, $C(x)$ is not convex.
	\begin{proof}
		Since $\alpha_n(x)$ is convex with a global minimum at $x=0$; if $k>0$, there exists a sufficiently large $x=K$ such that
		\begin{align}
			\min_n\alpha_n(K)>4,
		\end{align}
		hence,
		\begin{align}
			\mathbb{E}_{p_K}[4-\alpha_n(K)]<0,
		\end{align}
		which concludes the proof.
	\end{proof}
\end{proposition}

Hence, $C(x)$ is convex only when $k=0$, which is not meaningful. Fortunately, we are aiming for a quasi-convex loss instead of convex. Hence, we require $C''(x)\geq0$ whenever $|C'(x)|\leq \delta$ (for some $\delta>0$) from \autoref{thm:quasi}. Thus, we need
\begin{align}
	\mathbb{E}_{p_x}[4-\alpha_n(x)]\geq 0, &&\text{when }-\delta\leq Z_x\mathbb{E}_{p_x}[\alpha_n'(x)]\leq \delta.\label{eq:quasi}
\end{align}

\begin{lemma}\label{thm:cont}
	Our objective function $C(x)$ in \autoref{def:alpha} is quasi-convex in a bounded convex set if
	\begin{align*}
		\sum_{m,n}^{}p_x(m)p_x(n)e^{k(x_n-x_m)}\leq 4-\delta_0,
	\end{align*}
	for some small $\delta_0>0$,
	where $p_x(\cdot)$ is as in \autoref{def:p}.
	\begin{proof}
		Rearranging the equation, we get
		\begin{align}
			\sum_{m,n}^{}p_x(m)p_x(n)e^{k(x_n-x_m)}=&\mathbb{E}_{p_x}[e^{kx_n}]\mathbb{E}_{p_x}[e^{-kx_n}],\\
			=&\mathbb{E}_{p_x}[e^{k(x-x_n)}]\mathbb{E}_{p_x}[e^{-k(x-x_n)}],\nonumber
		\end{align}
		for any $x\in\Re$. If the condition holds, we have
		\begin{align}
			4-\delta_0\geq\mathbb{E}_{p_x}[e^{k(x-x_n)}]\mathbb{E}_{p_x}[e^{-k(x-x_n)}],\\
			16-4\delta_0\geq4\mathbb{E}_{p_x}[e^{k(x-x_n)}]\mathbb{E}_{p_x}[e^{-k(x-x_n)}].
		\end{align}
		$Z_x$ is bounded in a bounded convex set. Therefore, when $|Z_x\mathbb{E}_{p_x}[\alpha_n'(x)]|$ is bounded by $\delta$ as in \eqref{eq:quasi}, we get the following with a suitable $\delta$:
		\begin{align}
			16\geq&4\mathbb{E}_{p_x}[e^{k(x-x_n)}]\mathbb{E}_{p_x}[e^{-k(x-x_n)}]\\
			&+\left(\mathbb{E}_{p_x}[e^{k(x-x_n)}]-\mathbb{E}_{p_x}[e^{-k(x-x_n)}]\right)^2,\\
			\geq&\left(\mathbb{E}_{p_x}[e^{k(x-x_n)}]+\mathbb{E}_{p_x}[e^{-k(x-x_n)}]\right)^2.
		\end{align}
		Because $e^{k(x-x_n)}$ is always positive, we have
		\begin{align}
			4\geq&\left(\mathbb{E}_{p_x}[e^{k(x-x_n)}]+\mathbb{E}_{p_x}[e^{-k(x-x_n)}]\right),
		\end{align}
		which satisfies the quasi-convexity requirement in \eqref{eq:quasi} and concludes the proof.
	\end{proof}
\end{lemma}

\begin{lemma}\label{thm:F}
	The constraint function in \autoref{thm:cont}
	\begin{align}
		F(\boldsymbol{x})=\sum_{m,n}^{}p_x(m)p_x(n)e^{k(x_n-x_m)},
	\end{align}
	is convex in $\boldsymbol{x}=\{x_n\}_{n=1}^N$ when $\boldsymbol{p}=\{p_x(n)\}_{n=1}^N$ is free. 
	\begin{proof}
		Taking its derivatives, we have
		\begin{align}
			\dfrac{\delta F(\boldsymbol{x})}{\delta x_n}=&kp_x(n)\sum_{i\neq n}p_x(i)(e^{k(x_n-x_i)}-e^{k(x_i-x_n)}),\\
			\dfrac{\delta^2 F(\boldsymbol{x})}{\delta x_n^2}=&k^2p_x(n)\sum_{i\neq n}p_x(i)(e^{k(x_n-x_i)}+e^{k(x_i-x_n)}),\\
			\dfrac{\delta^2 F(\boldsymbol{x})}{\delta x_n\delta x_m}=&-k^2p_x(n)p_x(m)(e^{k(x_n-x_m)}+e^{k(x_m-x_n)}).
		\end{align}
		Since the second derivative is positive and the cross derivatives are negative, we observe that
		\begin{align}
			\left|\dfrac{\delta^2 F(\boldsymbol{x})}{\delta x_n^2}\right|=\sum_{m\neq n}^{}\left|\dfrac{\delta^2 F(\boldsymbol{x})}{\delta x_n\delta x_m}\right|.
		\end{align}
		Thus, the Hessian of $F(\boldsymbol{x})$ is diagonally dominant and consequently, positive semi-definite, i.e., $F(\boldsymbol{x})$ is convex with respect to $\boldsymbol{x}=\{x_n\}_{n=1}^N$. 
	\end{proof}
\end{lemma}

\begin{theorem}
	When $k$ is selected as
	\begin{align*}
		k=2.633,
	\end{align*}
	the objective function $C(\cdot)$ in \autoref{def:alpha} is quasi-convex for a bounded convex set.
	\begin{proof}
		Since $F(\boldsymbol{x})$ is convex from \autoref{thm:F}, it is maximum when the samples are at the boundaries, i.e., $x_n\in\{0,1\}$. Hence, we have
		\begin{align}
			F(\boldsymbol{x})\leq& q^2+(1-q)^2+q(1-q)(e^k+e^{-k}),\\
			\leq&1+q(1-q)(e^k+e^{-k}-2),
		\end{align}
		where $q$ is the sum of $p_x(n)$ for which $x_n$ is $0$. Thus,
		\begin{align}
			F(\boldsymbol{x})\leq&1+\frac{1}{4}(e^k+e^{-k}-2),
		\end{align}
		since $q(1-q)$ is concave and maximum at $q=1/2$. As per \autoref{thm:cont}, it suffices to choose a $k$ that makes this upper bound strictly less than $4$, i.e.,
		\begin{align}
			e^k+e^{-k}< 14
		\end{align}
		which is satisfied by
		\begin{align}
			k=2.633,
		\end{align}
		and concludes the proof.
	\end{proof}
\end{theorem}

\subsection{Derivative-Free Quasi-Convex Optimization}
Given a set of samples ${x}_1,\ldots, {x}_N \in\{0,1\}$; we have the following objective function
\begin{align}
	\min_{{x}\in\Re} C(x)
\end{align}
as in \autoref{def:alpha}. When $k=2.633$, we can acquire a quasi-convex loss function, which has the following properties.

\begin{remark}
	When $C(x)$ is quasi-convex, we have the following properties:
	\begin{itemize}
		\item $C(x)$ has a unique minimizer set $x^*\in [x_-,x_+]$. If $x_-=x_+$, it has a unique minimizer $x^*$. 
		\item For a set of points $x_1<x_2<x_3$, if $C(x_2)<\min(C(x_1),C(x_3))$; the minimizer $x^*\in(x_1,x_3)$.
		\item If $C(\cdot)$ is strictly quasi-convex and $C(x_1)=C(x_2)$ for some $x_1<x_2$, we have $x^*\in(x_1,x_2)$.
	\end{itemize}
\end{remark}

To solve a quasi-convex optimization problem using derivative-free methods, we can do the following. 

\begin{enumerate}
	\item At the beginning, sample the points $x=0$, $x=1/2$ and $x=1$ with their evaluations $C(0)$, $C(1/2)$ and $C(1)$. Set the sampled set $\mathcal{X}_S=\{0,1/2,1\}$.
	\item Let the minimizer set $\mathcal{X}_M$ be $\argmin_{x\in\mathcal{X}_S} C(x)$, where the operation $\argmin$ returns every point that minimizes the argument.\label{step:min}
	\item Let the potential set $\mathcal{X}_P$ be the union of $\mathcal{X}_M$ and its immediate left and right adjacent points (if exists) in the set $\mathcal{X}_S$
	\item Create the query set $\mathcal{X}_Q$ from the middle of every adjacent pair in $\mathcal{X}_P$. Hence $|\mathcal{X}_Q|=|\mathcal{X}_P|-1$. Set the sample set $\mathcal{X}_S=\mathcal{X}_P\bigcup\mathcal{X}_Q$.
	\item Return to Step \ref{step:min}.
\end{enumerate} 

\begin{remark}
	As a stopping criterion, we can utilize an $\epsilon$-closeness metric, where the algorithm stops whenever the adjacent samples in $\mathcal{X}_M$ has a distance less than $\epsilon$.
\end{remark}

\begin{remark}
	Note that the set $\mathcal{X}_M$ is an uninterrupted subset of $\mathcal{X}_S$, i.e., the samples correspond to adjacent samples from $\mathcal{X}_S$. In other words, if $x_1,x_2\in\mathcal{X}_M$, then
	\begin{align*}
		x\in\mathcal{X}_M, &&\forall x\in\mathcal{X}_S\cap[x_1,x_2].
	\end{align*}
\end{remark}

\begin{remark}
	If the objective function is strictly quasi-convex, we have two cases:
	\begin{enumerate}
		\item Either $|\mathcal{X}_M|=1$,
		\item Or $|\mathcal{X}_M|=2$. 
	\end{enumerate}
	
	We observe that when we are at the first scenario, we can go to either the first or the second scenario with two evaluations. However, if we are at the second scenario, we can only go to the first scenario with three evaluations.
	Hence, we can either halve the search space with two evaluations or quarter the search space in five evaluations. In either case the convergence to $\epsilon$ closeness is exponential, i.e., takes $O(N\log(\epsilon^{-1}))$ time.
\end{remark}

\begin{remark}
	If the function is quasi-convex but not strictly quasi-convex, the convergence is slower. In fact, in the regions lacking strict regularity, we have reciprocal convergence speed.
	At worse, we will have an evaluation overhead of $O(Np\epsilon^{-1})$, where $p$ is the the fraction of the isotonic regions. 
\end{remark}

\begin{remark}
	We can also be content with an approximate solution that is part of the sample set. In this case, complexity becomes $O(N\log(N))$.
\end{remark}

\section{Discussions and Conclusion}\label{sec:conc}
In literature, there are some desirable properties to have for centralizing metrics like the equivariance under scaling, translation, rotation or some other transform \cite{sarle1995measurement,drezner2002weber}. Nonetheless, it is straightforward to achieve equivariance with some preprocessing. Multivariate extension is also straightforward with a separate analysis in each dimension, which is reasonable since it is the case in absolute or square losses.

Although the L2-loss has fast learning performance, it is not robust against outliers. While the L1-loss has slower learning performance because of non-strict convexity, it is more robust. 
L0-loss provides higher robustness because of its bounded loss definition. All in all, an ideal loss function would be bounded for large errors and strictly convex for small errors.

To this end, we have proposed an extension to the generalized formulation of Huber loss. With this formulation, we achieve a smooth loss that is convex near $0$ for fast learning and concave for large $x$ for robustness. We show that by using the log-exp transform together with the logistic function; we can design a suitable loss that combines the desirable properties of L2-loss and L0-loss, i.e., bounded loss for large $x$ and strict convexity for small $x$.

Since the loss function is Lipschitz continuous, we show that with global optimization algorithms, it is possible to achieve a linear convergence rate. Moreover, with proper setting of the parameters, we prove that it is feasible to create a quasi-convex composite loss function. We propose a derivative-free algorithm that can find an optimal solution with exponential convergence speed.

\bibliographystyle{IEEEtran}
\bibliography{double_bib}

\begin{thebibliography}{10}
\providecommand{\url}[1]{#1}
\csname url@samestyle\endcsname
\providecommand{\newblock}{\relax}
\providecommand{\bibinfo}[2]{#2}
\providecommand{\BIBentrySTDinterwordspacing}{\spaceskip=0pt\relax}
\providecommand{\BIBentryALTinterwordstretchfactor}{4}
\providecommand{\BIBentryALTinterwordspacing}{\spaceskip=\fontdimen2\font plus
\BIBentryALTinterwordstretchfactor\fontdimen3\font minus
  \fontdimen4\font\relax}
\providecommand{\BIBforeignlanguage}[2]{{%
\expandafter\ifx\csname l@#1\endcsname\relax
\typeout{** WARNING: IEEEtran.bst: No hyphenation pattern has been}%
\typeout{** loaded for the language `#1'. Using the pattern for}%
\typeout{** the default language instead.}%
\else
\language=\csname l@#1\endcsname
\fi
#2}}
\providecommand{\BIBdecl}{\relax}
\BIBdecl

\bibitem{poor_book}
H.~V. Poor, \emph{An Introduction to Signal Detection and Estimation}.\hskip
  1em plus 0.5em minus 0.4em\relax NJ: Springer, 1994.

\bibitem{cesa_book}
N.~Cesa-Bianchi and G.~Lugosi, \emph{Prediction, learning, and games}.\hskip
  1em plus 0.5em minus 0.4em\relax Cambridge university press, 2006.

\bibitem{huberbook}
T.~Hastie, R.~Tibshirani, and J.~Friedman, \emph{The Elements of Statistical
  Learning}, ser. Springer Series in Statistics.\hskip 1em plus 0.5em minus
  0.4em\relax New York, NY, USA: Springer New York Inc., 2001.

\bibitem{portnoy2000robust}
S.~Portnoy and X.~He, ``A robust journey in the new millennium,'' \emph{Journal
  of the American Statistical Association}, vol.~95, no. 452, pp. 1331--1335,
  2000.

\bibitem{hastie2019statistical}
T.~Hastie, R.~Tibshirani, and M.~Wainwright, \emph{Statistical learning with
  sparsity: the lasso and generalizations}.\hskip 1em plus 0.5em minus
  0.4em\relax Chapman and Hall/CRC, 2019.

\bibitem{huber2004robust}
P.~J. Huber, \emph{Robust statistics}.\hskip 1em plus 0.5em minus 0.4em\relax
  John Wiley \& Sons, 2004, vol. 523.

\bibitem{barron2019general}
J.~T. Barron, ``A general and adaptive robust loss function,'' in
  \emph{Proceedings of the IEEE/CVF Conference on Computer Vision and Pattern
  Recognition}, 2019, pp. 4331--4339.

\bibitem{black1996unification}
M.~J. Black and A.~Rangarajan, ``On the unification of line processes, outlier
  rejection, and robust statistics with applications in early vision,''
  \emph{International journal of computer vision}, vol.~19, no.~1, pp. 57--91,
  1996.

\bibitem{zhang1997parameter}
Z.~Zhang, ``Parameter estimation techniques: A tutorial with application to
  conic fitting,'' \emph{Image and vision Computing}, vol.~15, no.~1, pp.
  59--76, 1997.

\bibitem{rosasco2004loss}
L.~Rosasco, E.~De~Vito, A.~Caponnetto, M.~Piana, and A.~Verri, ``Are loss
  functions all the same?'' \emph{Neural computation}, vol.~16, no.~5, pp.
  1063--1076, 2004.

\bibitem{hampel2011robust}
F.~R. Hampel, E.~M. Ronchetti, P.~J. Rousseeuw, and W.~A. Stahel, \emph{Robust
  statistics: the approach based on influence functions}.\hskip 1em plus 0.5em
  minus 0.4em\relax John Wiley \& Sons, 2011, vol. 196.

\bibitem{gokcesu2018density}
K.~Gokcesu and S.~S. Kozat, ``Online density estimation of nonstationary
  sources using exponential family of distributions,'' \emph{{IEEE} Trans.
  Neural Networks Learn. Syst.}, vol.~29, no.~9, pp. 4473--4478, 2018.

\bibitem{beck1977parameter}
J.~V. Beck and K.~J. Arnold, \emph{Parameter estimation in engineering and
  science}.\hskip 1em plus 0.5em minus 0.4em\relax James Beck, 1977.

\bibitem{gokcesu2016prediction}
N.~D. Vanli, K.~Gokcesu, M.~O. Sayin, H.~Yildiz, and S.~S. Kozat, ``Sequential
  prediction over hierarchical structures,'' \emph{IEEE Transactions on Signal
  Processing}, vol.~64, no.~23, pp. 6284--6298, Dec 2016.

\bibitem{singer}
A.~C. Singer and M.~Feder, ``Universal linear prediction by model order
  weighting,'' \emph{IEEE Transactions on Signal Processing}, vol.~47, no.~10,
  pp. 2685--2699, Oct 1999.

\bibitem{neyshabouri2018asymptotically}
M.~M. Neyshabouri, K.~Gokcesu, H.~Gokcesu, H.~Ozkan, and S.~S. Kozat,
  ``Asymptotically optimal contextual bandit algorithm using hierarchical
  structures,'' \emph{IEEE transactions on neural networks and learning
  systems}, vol.~30, no.~3, pp. 923--937, 2018.

\bibitem{rousseeuw2005robust}
P.~J. Rousseeuw and A.~M. Leroy, \emph{Robust regression and outlier
  detection}.\hskip 1em plus 0.5em minus 0.4em\relax John wiley \& sons, 2005.

\bibitem{gokcesu2018anomaly}
K.~Gokcesu and S.~S. Kozat, ``Online anomaly detection with minimax optimal
  density estimation in nonstationary environments,'' \emph{{IEEE} Trans.
  Signal Process.}, vol.~66, no.~5, pp. 1213--1227, 2018.

\bibitem{rousseeuw2011robust}
P.~J. Rousseeuw and M.~Hubert, ``Robust statistics for outlier detection,''
  \emph{Wiley interdisciplinary reviews: Data mining and knowledge discovery},
  vol.~1, no.~1, pp. 73--79, 2011.

\bibitem{gokcesu2019outlier}
K.~Gokcesu, M.~M. Neyshabouri, H.~Gokcesu, and S.~S. Kozat, ``Sequential
  outlier detection based on incremental decision trees,'' \emph{{IEEE} Trans.
  Signal Process.}, vol.~67, no.~4, pp. 993--1005, 2019.

\bibitem{lange1990convergence}
K.~Lange, ``Convergence of em image reconstruction algorithms with gibbs
  smoothing,'' \emph{IEEE transactions on medical imaging}, vol.~9, no.~4, pp.
  439--446, 1990.

\bibitem{charbonnier1997deterministic}
P.~Charbonnier, L.~Blanc-F{\'e}raud, G.~Aubert, and M.~Barlaud, ``Deterministic
  edge-preserving regularization in computed imaging,'' \emph{IEEE Transactions
  on image processing}, vol.~6, no.~2, pp. 298--311, 1997.

\bibitem{gokcesu2021generalized}
K.~Gokcesu and H.~Gokcesu, ``Generalized huber loss for robust learning and its
  efficient minimization for a robust statistics,'' \emph{arXiv preprint
  arXiv:2108.12627}, 2021.

\bibitem{logcosh}
R.~Neuneier and H.~G. Zimmermann, ``How to train neural networks,'' in
  \emph{Neural networks: tricks of the trade}.\hskip 1em plus 0.5em minus
  0.4em\relax Springer, 1998, pp. 373--423.

\bibitem{gokcesu2022low}
K.~Gokcesu and H.~Gokcesu, ``Low regret binary sampling method for efficient
  global optimization of univariate functions,'' \emph{arXiv preprint
  arXiv:2201.07164}, 2022.

\bibitem{sarle1995measurement}
W.~S. Sarle, ``Measurement theory: Frequently asked questions,''
  \emph{Disseminations of the International Statistical Applications
  Institute}, vol.~1, no.~4, pp. 61--66, 1995.

\bibitem{drezner2002weber}
Z.~Drezner, K.~Klamroth, A.~Sch{\"o}bel, and G.~O. Wesolowsky, ``The weber
  problem,'' \emph{Facility location: Applications and theory}, pp. 1--36,
  2002.

\end{thebibliography}

\end{document}